\documentclass[article, 10 pt, conference]{ieeeconf}  

\usepackage{graphics} 
\usepackage{graphicx}
\usepackage{subfig}
\usepackage{multirow}
\usepackage{rotating}
\usepackage{amsmath}

\usepackage{todonotes}

\usepackage[noadjust]{cite}

\newcommand{\subparagraph}{}
\usepackage{titlesec}
\titlespacing*{\section}{0pt}{*1}{*1}
\usepackage{subfig}
\usepackage{gensymb}
\usepackage[font=small]{caption}

\usepackage{setspace}

\title{\LARGE \bf
Learning to Take Good Pictures of People with a Robot Photographer 
}

\author{Rhys Newbury$^{1}$, Akansel Cosgun$^{1}$, Mehmet Koseoglu$^{2}$ and Tom Drummond$^{1}$\\
Australian Centre for Robotic Vision\\
$^{1}$Monash University, $^{2}$University of Melbourne}

\begin{document}

\maketitle
\thispagestyle{empty}
\pagestyle{empty}

\begin{abstract}
We present a robotic system capable of navigating autonomously by following a line and taking good quality pictures of people. When a group of people are detected, the robot rotates towards them and then back to line while continuously taking pictures from different angles. Each picture is processed in the cloud where its quality is estimated in a two-stage algorithm. First, features such as the face orientation and likelihood of facial emotions are input to a fully connected neural network to assign a quality score to each face. Second, a representation is extracted by abstracting faces from the image and it is input to a to Convolutional Neural Network (CNN) to classify the quality of the overall picture. We collected a dataset in which a picture was labeled as good quality if subjects are well-positioned in the image and oriented towards the camera with a pleasant expression. Our approach detected the quality of pictures with 78.4\% accuracy in this dataset, and received a better mean user rating (3.71/5) than a heuristic method that uses photographic composition procedures in a study where 97 human judges rated each picture. A statistical analysis against the state-of-the-art verified the quality of the resulting pictures.
\end{abstract}

\section{Introduction}
\label{sec:introduction}

The estimated value of digital photography market was $\$77.6$B in 2015~\cite{zion} and is expected to increase as social media use continues to grow. Photographs are typically generated by manually operating digital cameras and smartphones. Automating the picture taking process has great commercial potential. A robot photographer could be used for event photography, for example, to take pictures of attendees in a social event. We are interested in developing a robotic system capable of autonomously navigating and taking good pictures of people.

There are two algorithmic challenges in developing a robot photographer: how to position the robot base and how to identify whether the pictures taken are aesthetically pleasing or not. We adopt an approach often employed by professional photographers: take a bunch of pictures in various situations and choose the best ones afterwards. This design choice offloads the complexity to the evaluation of the image quality rather than the intelligent positioning of the robot base. Photographic composition rules such as the Rule of Thirds~\cite{peterson_2015} exist as a general guidance for taking good pictures. However, it is unclear how these hand-crafted rules can be combined into an algorithm to classify a picture as `good quality', especially when there are multiple subjects in the picture. In this work, we use deep learning techniques for automatic feature extraction and classification of the picture quality.

 The robotic system presented in this work navigates around the environment by following a tape plastered onto the floor by the end user. When a group of people is detected, the robot rotates towards them, takes a burst of pictures and uploads them to the cloud. Each image sent to the cloud is asynchronously processed as follows. First, new images are generated from the original image by applying a series of crops. This procedure results in slight zoom variations and increases the likelihood of capturing a better picture than the original. Second, faces are detected and features such as the angle of the face and emotions are extracted using Google Cloud Vision. Third, the quality of each face is scored using the extracted face features. Fourth, an abstract representation is generated from the face scores. Finally, a classifier is used to detect the quality of the overall picture. We evaluate our approach quantitatively by reporting classification accuracies and qualitatively with a user study.

 The organization of the paper is as follows. After reviewing the relevant literature in Sect.~\ref{sec:related_work}, we present the system overview in Sect.~\ref{sec:system_overview}. The picture taking behavior is explained in Sect.~\ref{sec:robotic_picture_taking}. The face quality classifier is described in Sect.~\ref{sec:detecting_good_face_photos}, followed by the overall picture  quality classifier in Sect.~\ref{sec:detecting_good_overall_photos}. The user study is presented in Sect.~\ref{sec:user_study}, before concluding in Sect.~\ref{sec:conclusion}.
 
\begin{figure}[t!]%
    \centering
    \subfloat{{\includegraphics[width=0.565\linewidth]{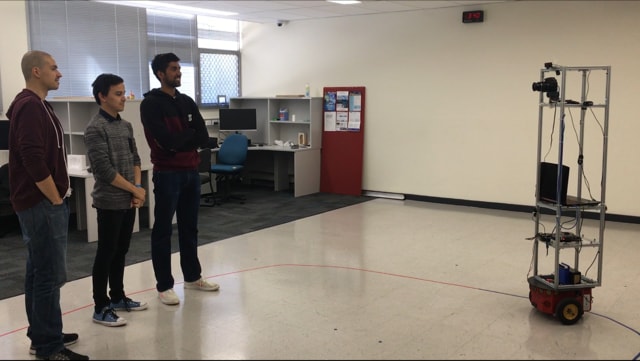}}}%
    \hspace{0.1em}%
    \subfloat{{\scalebox{-1}[1]{\includegraphics[width=0.43\linewidth]{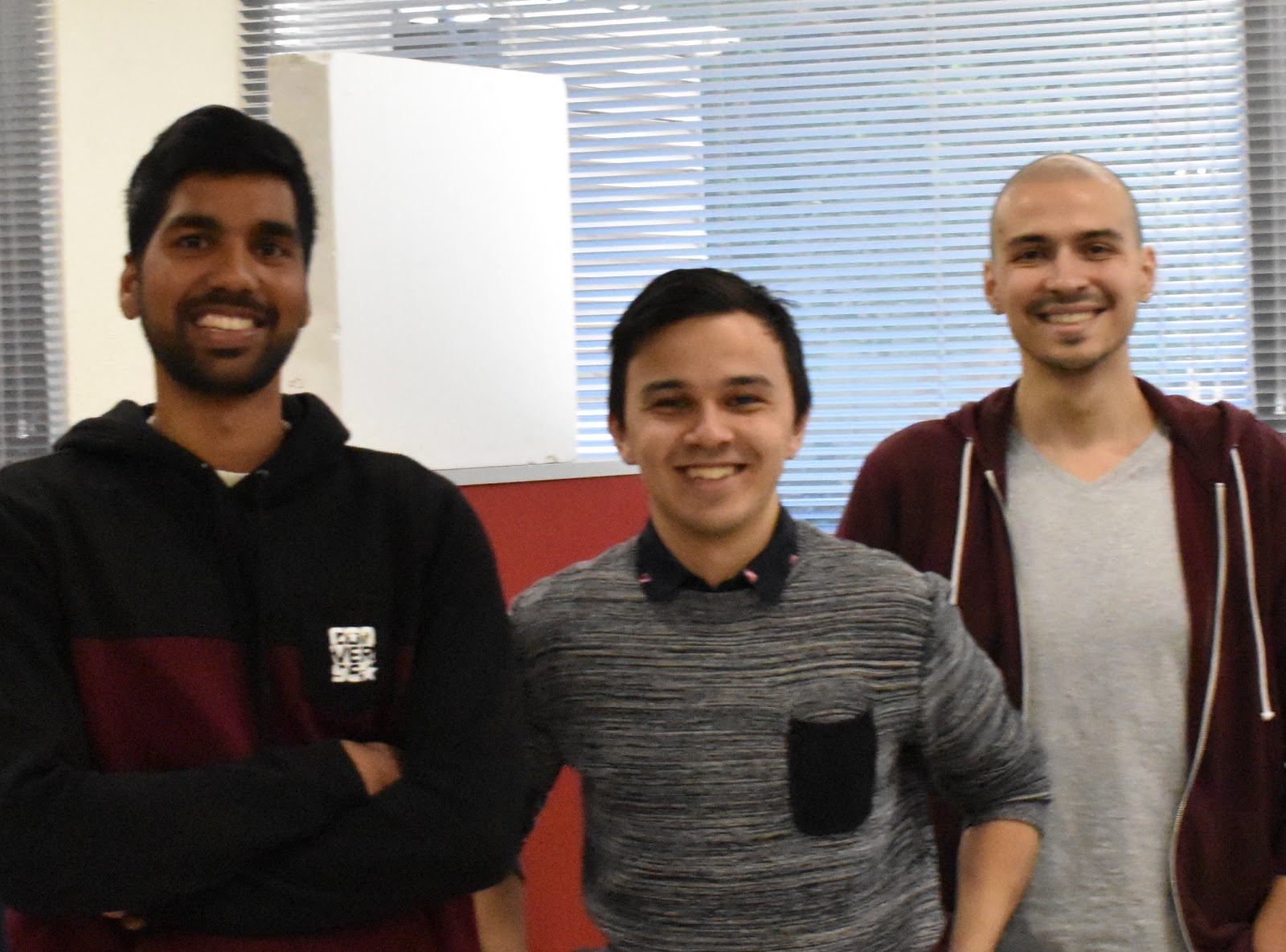}}}}%
    \caption{Left: The robot follows the tape on the floor and takes pictures of people. Right: A \textit{Good} quality picture taken by the robot.}%
    \label{fig:intro}
\end{figure}

\section{Related Work}
\label{sec:related_work}

We review the literature on various aspects of robotic photography including subject detection, navigation and photographic composition.

\subsection{Subject Detection}

Byers et al.~\cite{Byers} created the first robotic photography system using basic skin detection to detect subjects. Campbell and Pillai~\cite{Campbell} presented a robot photographer which found subjects based on measured motion parallax obtained via optical flow. The facial detection systems created by both Byers et al.\cite{Byers} and Campbell and Pillai~\cite{Campbell}, employed an ad hoc approach, however, it is almost impossible to detect faces in different poses with these methods. Furthermore, these techniques produced high amounts of false positives, where pictures were taken without any faces.\cite{face_detect}.

More recent approaches used different machine learning techniques to perform facial detection. Ahn et al.\cite{ Ahn} used Viola-Jones~\cite{viola2001rapid} which uses AdaBoost to leverage simple patterns to help detect faces. Zabaruskus and Cameron~\cite{zabarauskas2014luke} and Song et al.\cite{kim2010automatically} used variants of skin detection and additional heuristics to detect subjects. Luo et al.\cite{Luo}, Valenti et al.\cite{Valenti} and Fujimoto et al.~\cite{Fujimoto} used Haar features to perform subject detection. Hsu and Huang~\cite{Hsu} used DRMF (Discriminative Response Map Fitting) to detect faces which have a wide variety of orientations. 

These approaches do not leverage the success of deep learning in order to detect their subjects. The approach presented in our paper is similar to work by Lan and Sekiyama\cite{Lan2015, LAN2018} in which a convolutional neural network is used to detect the subjects faster and more accurately then previously possible.

\subsection{Robot Navigation}

A major problem in robotic photography is selecting a position in which the robot can take a nice picture. The main approaches include randomly wandering a room looking for photographic opportunities\cite{zabarauskas2014luke} and an objective function for moving to more favorable locations\cite{Byers}. The system presented by Hsu and Huang\cite{Hsu} navigated to a few predefined locations, however, the path of the robot was dynamically planned therefore making it harder for attendees to interpret the robot's motions. Predictable navigation in human environments has been addressed in previous work~\cite{cosgun2016anticipatory}.

The robotic system presented in our work follows a predefined path allowing the user complete control of how the robot will navigate its environment and allowing predictable behavior for bystanders.

\subsection{Photographic Composition}
\label{sec:photo_comp}

When aligning the subject in the frame many approaches~\cite{zabarauskas2014luke,Hsu,Ahn,Gadde,Valenti,Byers,Lan2015,Luo,Suzuki, Liu2010OptimizingPC,Cavalcanti} use a subset of the following rules:

\begin{itemize}
\setlength\itemsep{0pt}
\item Rule Of Thirds: Subject should be along the lines that divides the image into nine equal parts~\cite{peterson_2015}



\item Visual Balance: Visually salient objects are distributed evenly around the image~\cite{Liu2010OptimizingPC}


\item Golden Cross: Subject should be at the intersection of a vertical and horizontal line defined by the ratio 1.61


\item No Middle: The subject should not be in the middle of the image~\cite{zabarauskas2014luke}

\item No Edges: The edges of the frame should not pass through the subject~\cite{zabarauskas2014luke}

\item Occupancy: A third of the image should be taken up by subjects~\cite{zabarauskas2014luke}

\end{itemize}

Recent work in this area include using Kullback–Leibler divergence as a composition metric for a group of people~\cite{LAN2018}. Previous approaches are mostly suited to a single subject in images or do not consider additional features such as whether subjects are smiling or not.

In our approach, we learn the quality of the image from labeled images using machine learning instead of hand-crafted rules.

\section{System Overview}
\label{sec:system_overview}

\subsection{Hardware Configuration}
\label{sec:hardwareconfig}

\begin{figure}[t!]
\centering
\includegraphics[width=0.65\linewidth]{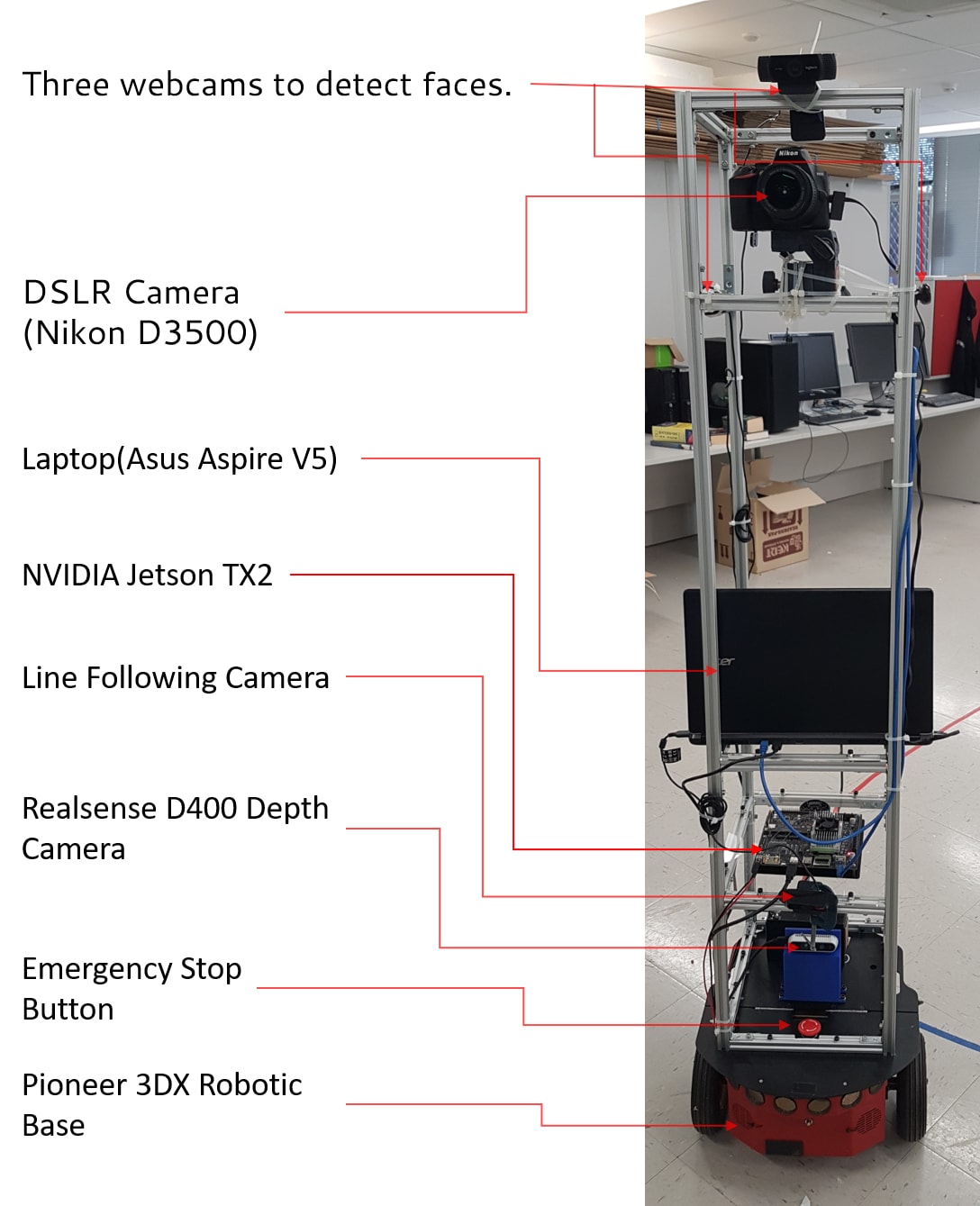}
\caption{Robot photographer}
\end{figure}

The system is built on the differential-drive Pioneer 3DX robotic base, which has a 23kg payload and top speed of 1.6m/s. Built on top of the Pioneer is a metal frame, allowing the cameras to be mounted at a height such that the angle of the picture is looking straight at the subject. The system consists of two computers (Aspire V5-552G and Nvidia Jetson TX2), four webcams, a Realsense D435 RGB-D camera and a Nikon D3500 DSLR camera. The TX2 has 8 GB RAM and an NVIDIA 256-core GPU, which allows it to run a neural network. The laptop has a quad core processor and 4GB of RAM. Attached to the TX2 are 3 webcams with a resolution of 640 by 480 pixels, which are running at a frame rate of 15fps. The output of the 3 webcams is used continuously for facial detection. The computer communicates to the robot using RS232 communication protocols. The TX2 is powered by a 12V lead acid battery, giving the computer around 12 hours of battery life. Attached to the laptop is another webcam similarly running at 15fps with a resolution of 640 by 480 pixels. This webcam is used to detect the line to allow the robot to navigate its environment. The Nikon D3500 is a 24.2 megapixel DSLR camera, which is connected to the computer via USB and uses \textit{libgphoto2} API for the open-source \textit{gPhoto} library to control the camera. Both computers use the Kinetic version of the Robot Operating System(ROS)\cite{ros} running on Ubuntu 16.04 LTS operating system.

\subsection{Architecture}
\label{sec:architecture}

\begin{figure}[ht!]
\centering
\includegraphics[clip, trim=3.75cm 6.25cm 8.6cm 4cm, width=0.5\textwidth]{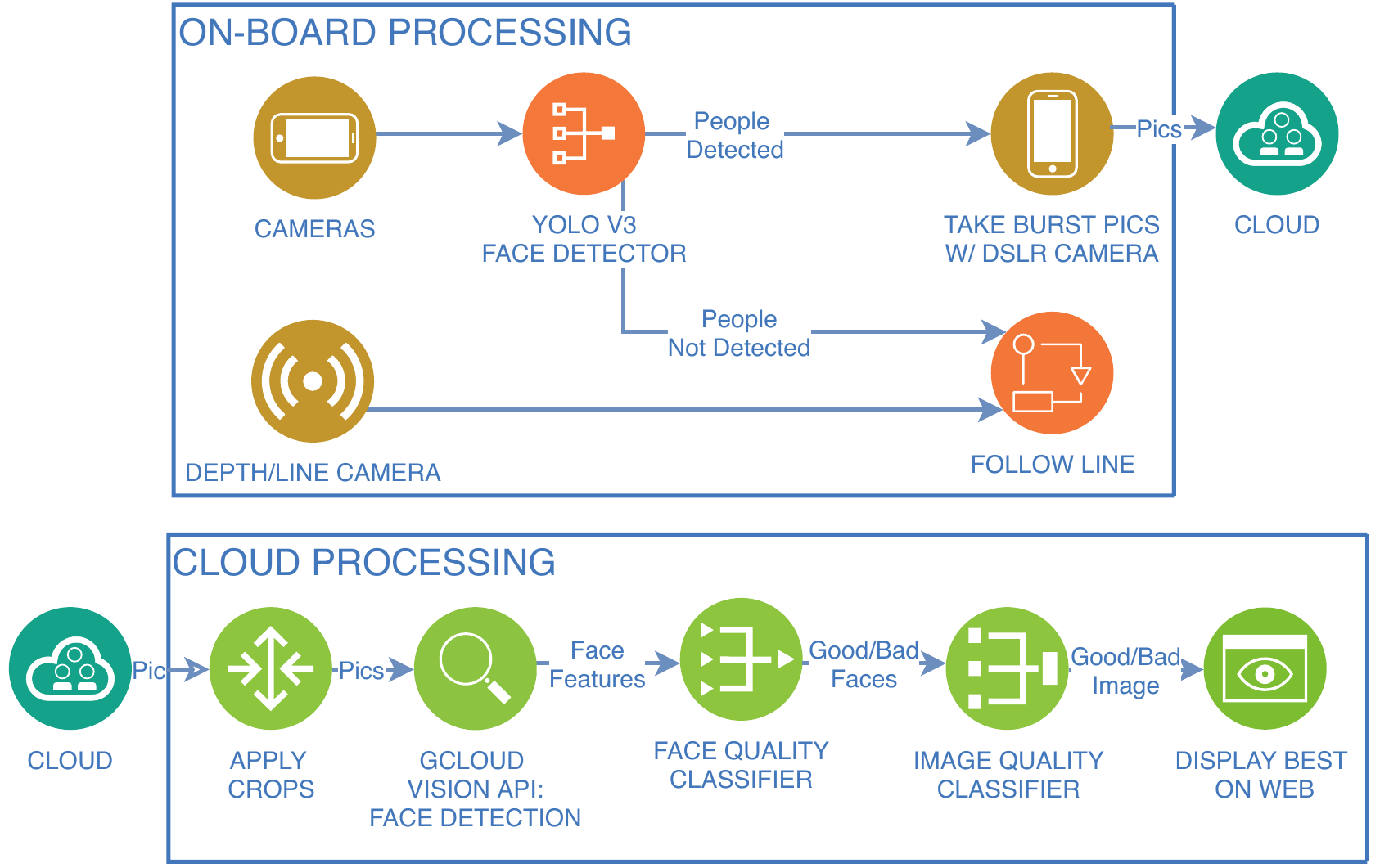}
\caption{System Architecture}
\label{fig:sys_architecture}
\end{figure}

The system architecture is shown in Figure~\ref{fig:sys_architecture}. The robot moves around the environment by following a line plotted by the user, as explained in Sect.~\ref{sec:navigation}. The collision prevention module which stops the robot for obstacles is described in Sect.~\ref{sec:collision_prevention}. While moving around the course, the robot performs face detection (Sect.~\ref{sec:face_detection}). When the robot sees an opportunity for a picture, it rotates towards the subject(s), and takes a series of burst pictures from different robot orientations with the DSLR camera. The robot then rotates back towards the line and continues line following behavior. The picture taking behavior is explained in Sect.~\ref{sec:picture_taking}.

The tasks related to post processing of the pictures are performed in the cloud. After applying a number of crops to each picture, we calculate a score for the quality of each face in the picture (Sect.~\ref{sec:detecting_good_face_photos}). The face scores are used to generate an abstract representation which is used to classify whether the overall picture is good or not (Sect.~\ref{sec:detecting_good_overall_photos}). The best pictures from the last ten bursts are displayed as a slideshow on a webpage.

\section{Taking Pictures with a Robot}
\label{sec:robotic_picture_taking}

For a robot to be able to autonomously acquire pictures of people, the robot should be able to navigate autonomously without colliding with obstacles, detect subjects and take potentially good pictures.

\subsection{Navigation}
\label{sec:navigation}

\begin{figure}[t!]%
    \centering
    \subfloat[]{\includegraphics[width=0.47\linewidth]{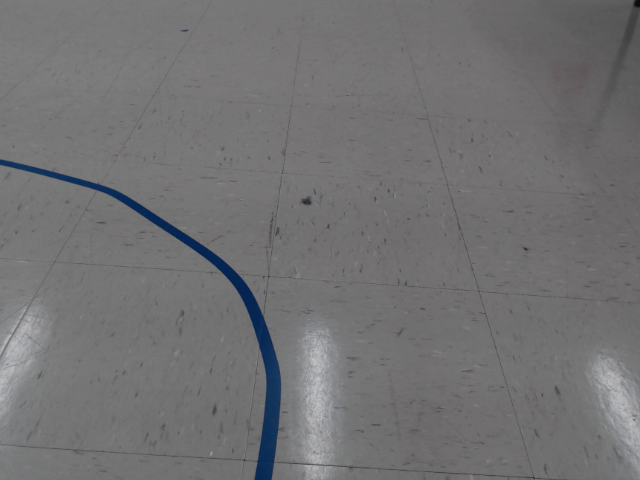} \label{line_a}} \hspace{1pt}
    \subfloat[]{\includegraphics[width=0.47\linewidth]{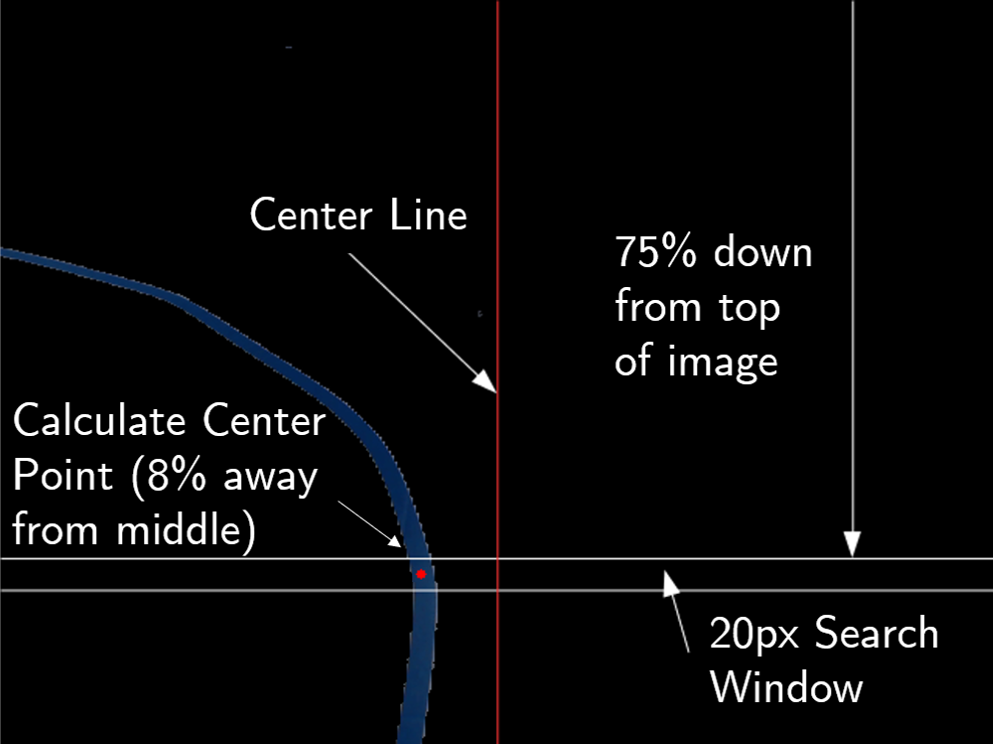} \label{line_b}} \hspace{1pt}
     \caption{a) Captured image from the line following camera. b) The Image after applying steps to caluclate centroid}
    \label{fig:user_study_qualitative}%
\end{figure}

We use line following as the main navigation method. An example image acquired by the line line following camera is shown in Figure \ref{line_a}. To distinguish a line from the rest of the image, a color mask is used to filter only the pixels which are colored similarly to the tape in the HSV color space. The upper and lower thresholds for the mask are found empirically. This approach, however, can create a noisy image if there are similarly colored objects in the image. Therefore, we only consider a vertical slice of 20 pixels, starting at 75\% down the image as shown in Figure \ref{line_b}. We calculate the image moments~\cite{OpenCV} using all the pixels in the vertical slice. An image moment is a weighted average of the pixel intensities and is computed by Equation \ref{eq:Moments}.

\begin{equation}
\begin{split}
I(x,y) =
\begin{cases}
0,& \text{if pixel value at (x,y) = 0} \\
1, &\text{otherwise}
\end{cases}
\label{eq:Moments2}
\end{split}
\end{equation}

\begin{equation}
m_{kl} = \sum_{x,y} I(x,y) \cdot (x)^k \cdot (y)^l
\label{eq:Moments}
\end{equation}

The center of the line in the horizontal direction is calculated through \(m_{01} / m_{00}\), illustrated as the red circle in Figure \ref{line_b}. A proportional controller is used to adjust the angular velocity, such that the calculated line centroid is in the middle of the image. Robot moves with a constant linear velocity as long as there are no obstacles on its path. 

We chose line following, a rather simple navigation model, over planning based methods because of two reasons. First, bystanders would feel more comfortable around the robot the robot's motions are predictable. Second, mapping and localization in crowded environments is very challenging.~\cite{morioka2011vision}.

\subsection{Collision Prevention}
\label{sec:collision_prevention}

We use a RGB-D camera to detect and stop for the objects in front of the robot. The robot does not deviate from the line but rather wait for the blocking dynamic obstacle (i.e. people) to move. We first convert the point cloud from the camera to a pseudo laser scan using the ROS package \texttt{\detokenize{depth_image_to_laserscan}}. This module takes a vertical slice of the depth image and projects it to a horizontal plane in the world frame. The distance values reported by the laser scan are then converted to cartesian coordinates in the robot frame, where the \textit{z-axis} is pointing away from the sensor and the \textit{x-axis} is pointing to the right of the robot. Any points detected within the robot footprint are ignored. This allows discarding the points coming from the metal frame in front of the camera. If there are at least $n_{stop}$ points in the laser scan such that $x<x_{stop}$ and $z<z_{stop}$ for $n_{detected}$ out of $n_{window}$ consecutive frames, the robot stops moving. The robot can only continue moving \(T_{stop}\) seconds after the object is cleared. The following parameters were used: $n_{stop}=10, x_{stop}=0.5m, z_{stop} = 2m , n_{detected} = 4 , n_{window} = 5, T_{stop} = 2s$.

Due to sensor noise, the robot occasionally detects obstacles when there are none. We deliberately made the obstacle detection oversensitive since this ensures that the robot stops for people and moving objects. The robot did not collide with any objects or persons during experiments.

\subsection{On-board Face Detection}
\label{sec:face_detection}

Our system is designed to only take pictures of people, thus the detection of faces is a critical component. We use YOLO V3\cite{yolo}, a deep learning based classifier, for on-board real-time face detection. We train YOLOv3\cite{yolo} on the WIDER Face dataset\cite{yang2016wider}, which contains $393$k face images with a high degree of variability in size, poses and occlusions. We use the three cameras placed on the metal frame allowing the robot to have close to a $180\degree$ field of view. The three images are joined horizontally, to form a single image, which is run through the neural network for face detection. Running the detector on the joint image was practically faster then running the neural network separately for each one of the images and combining the results.

\subsection{Picture Taking}
\label{sec:picture_taking}

Since the robot is intended to be used in social settings, our approach favors taking pictures of groups of people. While the robot is navigating, the id of the camera stream which has detected the most number of faces is stored for the previous 10 frames. If a camera stream has the most number of faces for $n_{max}$ out of the previous $n_{window}$ frames, then the robot rotates a fixed amount of degrees depending on the camera. If the chosen camera is the left camera it rotates $\theta_{side}$ degrees counter clockwise, if the chosen camera is the right camera it rotates $\theta_{side}$ degrees clockwise, if the chosen camera is the front camera, it rotates $\theta_{front}$ degrees clockwise. The robot then starts taking $n_{burst}$ burst pictures while rotating back towards the line. Once the line is detected, the robot continues line following behavior as described in Sect.~\ref{sec:navigation}. The follow parameters used: $n_{max}=7, n_{window}=10, \theta_{side}=130, \theta_{front}=40, n_{burst}=20$.
The taken pictures are moved from the camera memory to the laptop, which takes around 5 seconds. During this time the robot continues line following. This ensures the robot doesn't repeatedly take pictures from the same position. The pictures taken are uploaded to the cloud which are then cropped 6 times, reducing the original image by 600 by 400 pixels every iteration analogous to performing digital zoom. Each obtained picture is then cropped again using the Google Cloud Vision API Crop Hints which crops the frame around the dominant object while decreasing the aspect ratio from 1.5 to 1.33.


\section{Detecting Good Quality Faces in Pictures}
\label{sec:detecting_good_face_photos}

In this section we present two face quality detection algorithms, the data collection procedure and classification performance.

\subsection{Algorithms}

For each image obtained through the procedure described in Sect.~\ref{sec:picture_taking}, we use the Google Cloud Vision API to detect the faces. We propose two approaches to detect the quality of each face: Face CNN which uses the raw face image as input and Face Artificial Neural Network(ANN) which uses a set of face features as input.

\subsubsection{Face CNN}
\label{FaceCNN}
A CNN (Convolutional Neural Network)~\cite{krizhevsky2012imagenet} was trained to classify if a detected face is of good or bad quality. The input to the network is the raw image of the face. The input image is scaled to 40 by 30 pixels and converted to a grayscale image. The network structure includes 5 convolutional layers all with a filter size of (3,3), a stride of 2, and using the ReLU activation. The convolutional layers have 96,96,96,192,192 channels, respectively. Connected to the output of the convolutional layers are 7 fully connected layers with 100,200,400,800,400,200 and 10 nodes respectively. Finally, there is one output layer with sigmoid activation where the output represents the quality of the face image.

\subsubsection{Face ANN}
\label{FaceANN}

A fully-connected ANN (Artificial Neural Network) was trained to classify if a detected face is of good or bad quality. The input to the network is $9$ continuous numbers, which is a subset of the features outputted by the face detection function in Google Cloud Vision, listed below:
\begin{itemize}
\setlength\itemsep{0pt}
    \item Face orientation (roll, pitch, yaw)
    \item Likelihood of facial emotions (joy, sorrow, anger, surprise)
    \item Image features (exposure and blurriness)
\end{itemize}
The network has 5 hidden fully connected layers (32,64,64,32,16), all using ReLU Activation. Finally, there is one output layer, with a sigmoid activation where the output represents the quality of the face image.

\subsection{Data Collection}

We gathered a face image dataset from a combination of the lab experiments and an external dataset. We ran face detection on the 9917 pictures gathered during the development of the robot around the lab and added a further 3697 face images from the WIDER Face Data set\cite{yang2016wider}. We hand labelled each of the face images either as a good quality or a bad quality. A face is labelled as good quality if it is well exposed, in focus and the subject is looking at the camera with a pleasant expression. Otherwise, it is labelled as a bad quality face. Pictures that contain no faces and images that were smaller than $30$x$30$ were discarded. The face features obtained from Google Cloud Vision were also stored. A total of 6348 labeled face images were randomly split into training (80\%), test (10\%) and validation (10\%) sets.


\subsection{Classification Results}
As shown in Table~\ref{lab:table1}, the mean recognition accuracy over was 92.7\% for Face ANN and 83.2\% for Face CNN. Accuracy over three runs is averaged to reach these numbers.

\begin{table}[h!]
\centering
\begin{tabular}{c|c}
Method & Mean Accuracy \\ \hline
Face CNN & 83.2\%           \\
Face ANN & 92.7\%          
\end{tabular}
\caption{Recognition accuracy of good quality face classifiers. Input for Face CNN is the raw face image and the face feature vector for Face ANN}
\label{lab:table1}
\end{table}

The Face ANN performed significantly better than Face CNN despite having a much smaller network architecture. Abstraction and descriptive features likely boosted the Face ANN performance, compared to Face CNN which uses the raw face image. We can claim that features such as face orientation and facial emotions are descriptive for distinguishing good and bad quality face pictures. We believe that features such as the likelihood of closed eyes can further improve the performance. It is possible that Face CNN would perform better with a larger dataset.

The 92.7\% recognition accuracy for face quality is promising for this application, therefore we utilize the estimated face quality scores in overall picture quality detection.



\section{Detecting Good Quality Overall Pictures}
\label{sec:detecting_good_overall_photos}

In Sect.\ref{sec:detecting_good_face_photos} we estimate the quality of each face in isolation however the position and size of the faces in the image are also important in determining overall picture quality. For example, if a good quality face is detected on the very edge of a picture, it is most likely a bad picture overall. In this section we propose three approaches which use the quality, pixel size and location of each face in order to detect the quality of each picture. We also describe the data collection process and report the classifier performance. 

\subsection{Algorithms}
\label{sec:pic_quality_algorithms}
We present three approach for good quality picture detection: baseline which optimizes thresholds so that a couple of compositional rules are adhered, heuristic that optimizes for compositional rules as well as the face quality, and Picture CNN which uses supervised machine learning.

\subsubsection{Baseline}
\label{Baseline}

We describe a baseline algorithm that uses a variation of the photographic rules No Edge and Occupancy, which are described in Sect.\ref{sec:photo_comp}. The result of the face detection is a set of bounding boxes given the input image. A bounding box around a face is defined as two tuples $(x_{tl},y_{tl})$ and $(x_{br},y_{br})$, which are the coordinates for the top left and bottom right pixels. We check whether each face position satisfies the following constraints:
\begin{equation} 
\begin{split}
\label{eq1}
\hspace{-2cm}\frac{x_{tl}}{2x_c}>x_{min},
\,\frac{x_{br}}{2x_c}<x_{max},\,
\frac{y_{tl}}{2y_c}>y_{min},\,
\frac{y_{br}}{2y_c}<y_{max},\\
&\hspace{-8cm} occ_{min}<\frac{|x_{br}-x_{tl}||y_{br}-y_{tl}|}{2x_c\cdot 2y_c}<occ_{max}
\end{split}
\end{equation}
where $x_{min},x_{max},y_{min},y_{max}$ are thresholds for the normalized position of the face, $occ_{min},occ_{max}$ are the thresholds for the area occupancy of the face and $(x_c,y_c)$ are the coordinates of the center pixel in the image. If any of the faces in a picture does not satisfy any of the inequalities in Equation~\ref{eq1}, or there are no faces detected in the image, then the picture is automatically scored as zero. Otherwise, it is deemed a good picture and a non-zero score is calculated. The six thresholds were optimized such that the maximum detection accuracy on the training set is achieved. A genetic algorithm is used for optimization. For each picture labeled as labelled as good quality, we calculate a score as follows. First, the center pixel of a detected face $j$ in an image $i$ is found as:
\begin{equation}
\label{eq:baseline1}
\begin{split}
(x^i_{j_c}, y^i_{j_c}) =\left(\frac{x^i_{j_{br}} + x^i_{j_{tl}}}{2}, \frac{y^i_{j_{br}} + y^i_{j_{tl}}}{2}\right)
\end{split}
\end{equation}
The normalized distance $d_j^i$ from the image center $(x^i_c, y^i_c)$ for each face $j$ in image $i$ is then found by:
\begin{equation}
\label{eq:baseline2}
\begin{split}
d_j^i = \frac{\sqrt{(x^i_{j_c} - x^i_c)^2+(y^i_{j_c} - y^i_c)^2}}{\sqrt{(x^i_c)^2+(y^i_c)^2}}
\end{split}
\end{equation}
Finally, the overall score $S_i$ for image $i$ is computed as: 
\begin{equation}
\label{eq:baseline3}
\begin{split}
S_i=\sum_{j=0}^{j=n_{faces}}(1-d_j^i)\end{split}
\end{equation}
The score $S_i$ is used in the selection algorithm for choosing the best pictures, as discussed in Sect.~\ref{sec:user_study_photo_selection}.

\subsubsection{Heuristic}
The Heuristic approach is similar to the Baseline, except it also utilizes the face quality score. Two new thresholds are introduced regarding face quality: $r_{min}$ which is the minimum score for each face to be classified as good and $p_{min}$ which is the minimum proportion of faces that must be classified as good among all of the faces in the image. The additional constraint is below:
\begin{equation}
\label{eq:heuristic}
\begin{split}
b_j^i =
\begin{cases}
1,& \text{if}\:r_j^i > r_{min} \\
0, &\text{otherwise}
\end{cases}
\end{split}
\end{equation}
\begin{equation}
\label{eq:heuristic}
\begin{split}
\frac{\sum_{j=0}^{j=n_{faces}}\:b_j^i}{n_{faces}} > p_{min}
\end{split}
\end{equation}
where $r_j^i$ is the rating for face $j$ in image $i$. If any of the faces in a picture fails to satisfy any of the inequalities in Equation~\ref{eq1} or Equation~\ref{eq:heuristic}, or there are no faces detected in the image, then the picture is automatically scored as zero. Otherwise, it is deemed a good picture and a non-zero score is calculated. The eight thresholds were optimized using genetic algorithms such that the maximum detection accuracy on the training set is yielded. The overall score $S_i$ for image $i$ is found as: 
\begin{equation}
\label{eq:heuristic2}
\begin{split}
S_i = \sum_{j=0}^{j=n_{faces}}(1-d_j^i) \cdot r_j^i\end{split}
\end{equation}
The score $S_i$ is used in the selection algorithm for choosing the best pictures, as discussed in Sect.~\ref{sec:user_study_photo_selection}.

\subsubsection{Picture CNN}
\label{CNN_apppraoch}

The Heuristic and Baseline methods described in previous sections considered each face separately to determine the image quality. However, the relative sizes and positions of the faces in an image also contribute to the quality of the image. Machine learning techniques can be used to extract meaningful features that contribute to the quality of an image.

An abstract representation is generated from the face positions, size in pixels and scores. We use the face quality scores generated by Face ANN (Sect.~\ref{FaceANN}). The abstract representation is a grayscale image with the same resolution as the input image. It consists of a white background with gray rectangles in the corresponding locations of the face bounding boxes in the original image. This representation is passed to a CNN which extracts spatial relationships between faces. This approach will be referred to as Picture CNN for the rest of the paper. For an image $i$, the gray intensity $G_j^i$ of each rectangle is based on the corresponding quality score $r^i_j$ of face $j$, as follows:
\begin{equation}
G_j^i = 245 * r^i_j
\end{equation}
The maximum value 245 instead of 255 was chosen as it provides separation between the white background and the colored rectangles.

The neural network architecture has 2 convolutional layers all with a filter size of (4,4), a stride of 3, and using the LeakyReLU activation. The convolutional layers have 8 and 20 channels, respectively. Connected to the output of the convolutional layers are 2 fully connected layers with 1260 and 100 nodes respectively. Finally, there is one output layer with sigmoid activation where the output represents the quality of the input image.

\subsection{Data Collection}
We collected data in three different locations with volunteer subjects. The robot operated in autonomous picture taking behavior as described in Sect.~\ref{sec:robotic_picture_taking}. A total of 6580 pictures were collected after pictures without people in them were discarded. The collected images were hand labeled as one of the two classes: good or bad, with respect to their aesthetic quality. The pictures were randomly split into training (80\%), test (10\%) and validation (10\%) sets.

\subsection{Classification Results}
 
Table \ref{lab:table2} shows the mean recognition accuracy of the good quality picture detection methods and the face quality method they used.
 
\begin{table}[h!]
\centering
\begin{tabular}{l|c|c}
Pic Quality & Face Quality & Accuracy \\ \hline
Baseline  & None & 68.0\%           \\
Heuristic   & Face ANN  & 76.7\%           \\
CNN & None & 73.6\%           \\ 
Picture CNN & Face ANN     & 78.4\%           \\ 
\end{tabular}
\caption{Mean recognition accuracy for detecting good quality overall pictures}
\label{lab:table2}
\end{table}
\setlength{\tabcolsep}{0.15cm}
 
Using features extracted with deep learning techniques performed better than hand-crafted photographic composition rules (even with optimal parameters) in detecting good quality pictures. Among the three algorithms presented in Sect.~\ref{sec:pic_quality_algorithms}, Picture CNN has the highest recognition accuracy with 78.4\%, followed with the Heuristic method with 76.7\% and Baseline with 68.0\%. 

A CNN trained from the raw images rather than the abstract representation achieved 73.6\% recognition accuracy, which is worse than Picture CNN. We think that the Picture CNN generalized better because the input representation compactly encodes what determines the picture's quality: positions, scores and sizes of the faces.

Heuristic had a higher performance than the Baseline. Considering that the difference between the two methods is the use of face scores in the optimization, we can say that utilizing face quality improves the recognition accuracy of detecting good overall pictures.


We analyzed how the three methods perform with respect to the number of people in the pictures, see Table~\ref{lab:table3}. Picture CNN performs the best with more than 87.7\% accuracy when there are one or two faces in the image. Moreover, the accuracy for all three methods decreased when there are three or more faces in the picture. We attribute it to most pictures in our dataset having either one or two faces. The accuracy would improve with more data, especially if the pictures include three or more faces in them.

\begin{table}[t!]
\centering
\begin{tabular}{l|c|c|c}
Method    & 1 Face  & 2 Faces & 3+ Faces \\ \hline
Baseline  & 67.7\% & 71.7\% & 57.1\%  \\
Heuristic & 83.4\% & 68.1\%  & 66.8\%  \\
Picture CNN & 87.7\% & 87.8\% & 72.9\% 
\end{tabular}
\caption{Recognition accuracy for different number of faces in the picture}
\label{lab:table3}
\end{table}




We illustrate the strengths and limits of the Picture CNN approach with four classification shown in Table~\ref{tab:gt}. In the top left picture, all three subjects are smiling and positioned well, and the picture is classified correctly as \textit{Good}. In the bottom left picture, the subjects are the same and similarly positioned. One of the subjects (rightmost), however, is not smiling which led the algorithm to incorrectly classify the image as as \textit{Bad}. This example shows how our approach puts an importance to smiling, sometimes to a fault. The bottom right picture is an obviously \textit{Bad} picture. However, the face detector can not detect the subject who is bending over. Picture CNN classifies the picture as \textit{Good}, since it considers the only two subjects who are positioned well and is smiling. The same problem occurs when the face detector fails, for example, when faces are cut off from the picture. Detecting people and not just faces would alleviate this problem.




\setlength{\tabcolsep}{0.5mm}
\begin{table}[ht!]
\centering
\begin{tabular}{|c|c|c|c|}
\cline{3-4}
\multicolumn{2}{c}{} & \multicolumn{2}{|c|}{Correct Label} \\
\cline{3-4}
\multicolumn{2}{r|}{} & Good & Bad \\
\hline
\multirow{2}{*}{\rotatebox[origin=r]{90}{Picture CNN Output}} & 
\raisebox{1cm}{\rotatebox{90}{Good}} & \raisebox{-0.7mm}{\includegraphics[width=0.45\linewidth]{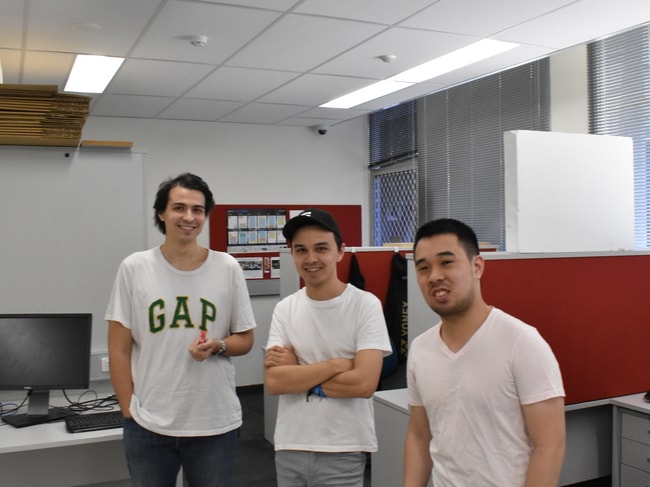}} & \raisebox{-0.7mm}{\includegraphics[width=0.45\linewidth]{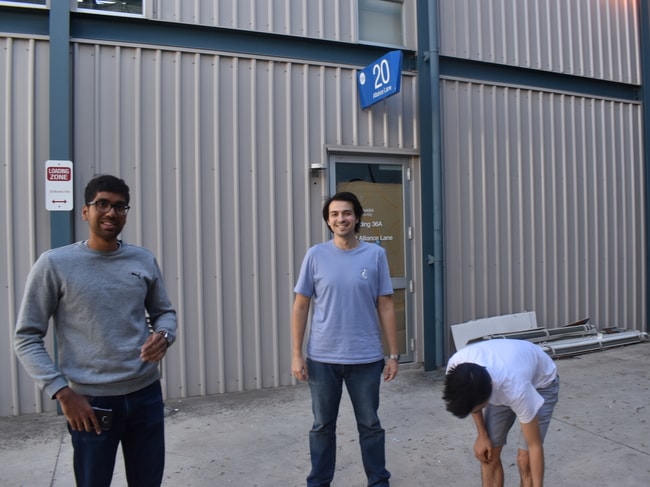}} \\ 
\cline{2-4} & \raisebox{1cm}{\rotatebox{90}{Bad}} & \raisebox{-0.7mm}{\includegraphics[width=0.45\linewidth]{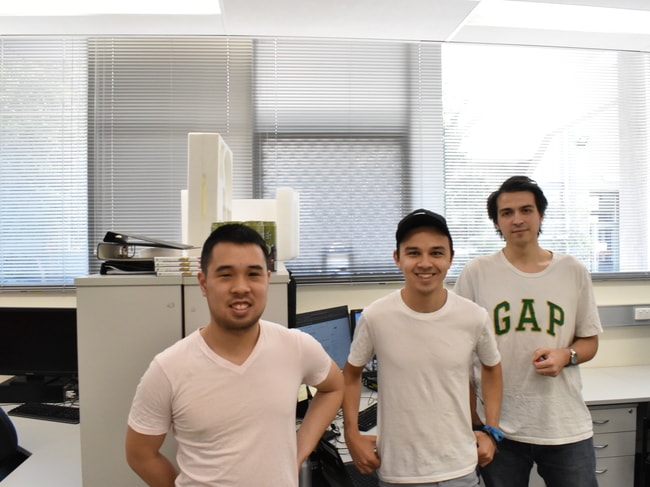}} & 
\raisebox{-0.7mm}{\includegraphics[width=0.45\linewidth]{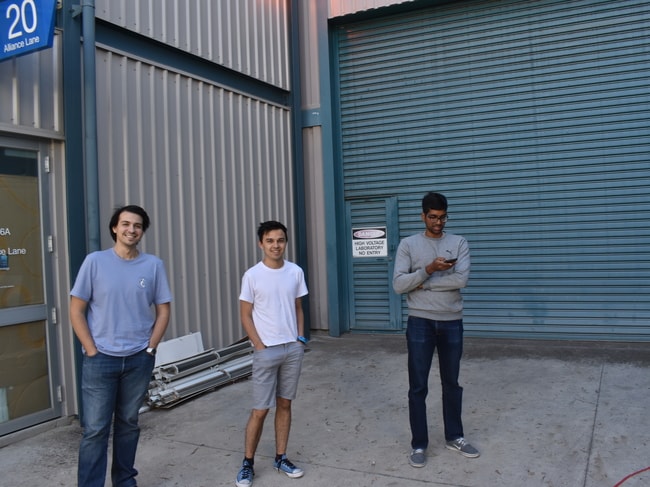}} \\ \hline
\end{tabular}
\caption{Picture CNN classification examples}
\label{tab:gt}
\end{table}
\setlength{\tabcolsep}{0.15cm}


\section{User Study}
\label{sec:user_study}

We conducted a user study in order to measure how the pictures chosen by the Picture CNN method would be judged against the Baseline and Heuristic methods.

A new set of pictures were collected for the user study. These pictures included new volunteers and locations that were not included in the original dataset used for training the neural network models. All the burst pictures, along with randomly cropped versions were fed to each of the Baseline, Heuristic and Picture CNN methods for classification. In the next section we describe the algorithm used for selecting the best pictures for each method.

\subsection{Picture Selection}
\label{sec:user_study_photo_selection}

Among all the pictures taken, we select 24 pictures for each method. The objective is to select pictures that are different from each other, just like a human photographer would. Rather than the detected label, we used the detection score that is a continuous number between 0 and 1 for each classifier. Sorting with respect to this score and selecting the top 24 would result in very similar pictures being chosen. For example, it is likely for a cropped version of a good picture to also receive high scores. Same is true for pictures taken in the same burst. We employ two constraints in order to introduce variation in the selected pictures and allow the subjects to evaluate the effectiveness of the algorithm in different situations:

\begin{itemize}
\setlength\itemsep{0pt}
\item Variety in the number of people: We constrained the selected images to have an equal number (8 each) of pictures for 1-face, 2-face and 3+ face pictures.
\item One picture from each burst: For each method we allow a single picture from each burst. 
\end{itemize}


The order in which face category would be selected first would change the selected pictures since we only allow a single picture from each burst. In the dataset collected for the user study, the number of 1-Face pictures was the most, followed by 2-Faces and 3+ Faces. The selection algorithm was executed in the reverse order, to allow the underrepresented categories to select the best pictures in a smaller number of images.

10 of the selected images were identical for Heuristic and Baseline, therefore the user study consisted of a total of 62 pictures. 

\subsection{Design}
\label{sec:user_study_design}

The selected pictures were posted to an online survey using Google Forms, where 100 subjects were asked to evaluate the quality of the 62 pictures. The subjects were recruited using personal connections and were different than people who volunteered to be in the photographs. The subjects completed the survey online and were not compensated for their effort. The subjects were asked to rate the quality of the pictures considering the positioning of the subjects and disregard the quality of the camera itself. A 5-point Likert scale where the subjects rated each pictures with one of the following: \textit{(1) Very bad, (2) Bad, (3) Neutral, (4) Good and (5) Very Good}. The order of the images were randomized for each subject.

\subsection{Hypotheses}
\label{sec:user_study_measures}
We pose two research hypotheses:

\textit{H1:} Picture CNN will outperform Heuristic and Baseline

\textit{H2:} Picture CNN will outperform the state-of-the-art~\cite{zabarauskas2014luke,Ahn,Byers}

\subsection{Results}
\label{sec:user_study_result}

We first removed responses where a subject gave the same rating to every picture. This led to removal of three responses in which the subjects rated all pictures \textit{(5) Very Good}. We base our analysis on the remaining 97 subject responses. Each subject rated 24 pictures of each method, therefore we received 2328 ratings for each method. We convert the Likert scale ratings to corresponding numerical values between $1$ to $5$. This is not the most proper way to treat ordinal data, however it makes our analysis straightforward and the results easy to interpret. The mean and standard deviation of the ratings are shown in Table \ref{tab:user_study_method_comparison}.

\begin{table}[h!]
\centering
\begin{tabular}{l|c|c}
Method    & $\mu$  & $\sigma$ \\ \hline
Baseline  & 3.58 & 1.06 \\
Heuristic & 3.63 & 1.04  \\
Picture CNN & 3.71 & 1.06 
\end{tabular}
\caption{Mean and standard deviation of user ratings}
\label{tab:user_study_method_comparison}
\end{table}

Picture CNN has the best average rating with $3.71$, followed by Heuristic with $3.63$. As expected, Baseline performed the worst with $3.58$. The standard deviations of ratings were roughly $1$ for all three methods. Pictures with that received the best, worst and around median mean ratings are shown in Fig. \ref{fig:user_study_qualitative}.


\begin{figure}[t!]%
    \centering
    \subfloat[Mean rating: $4.25$]{\includegraphics[width=0.32\linewidth]{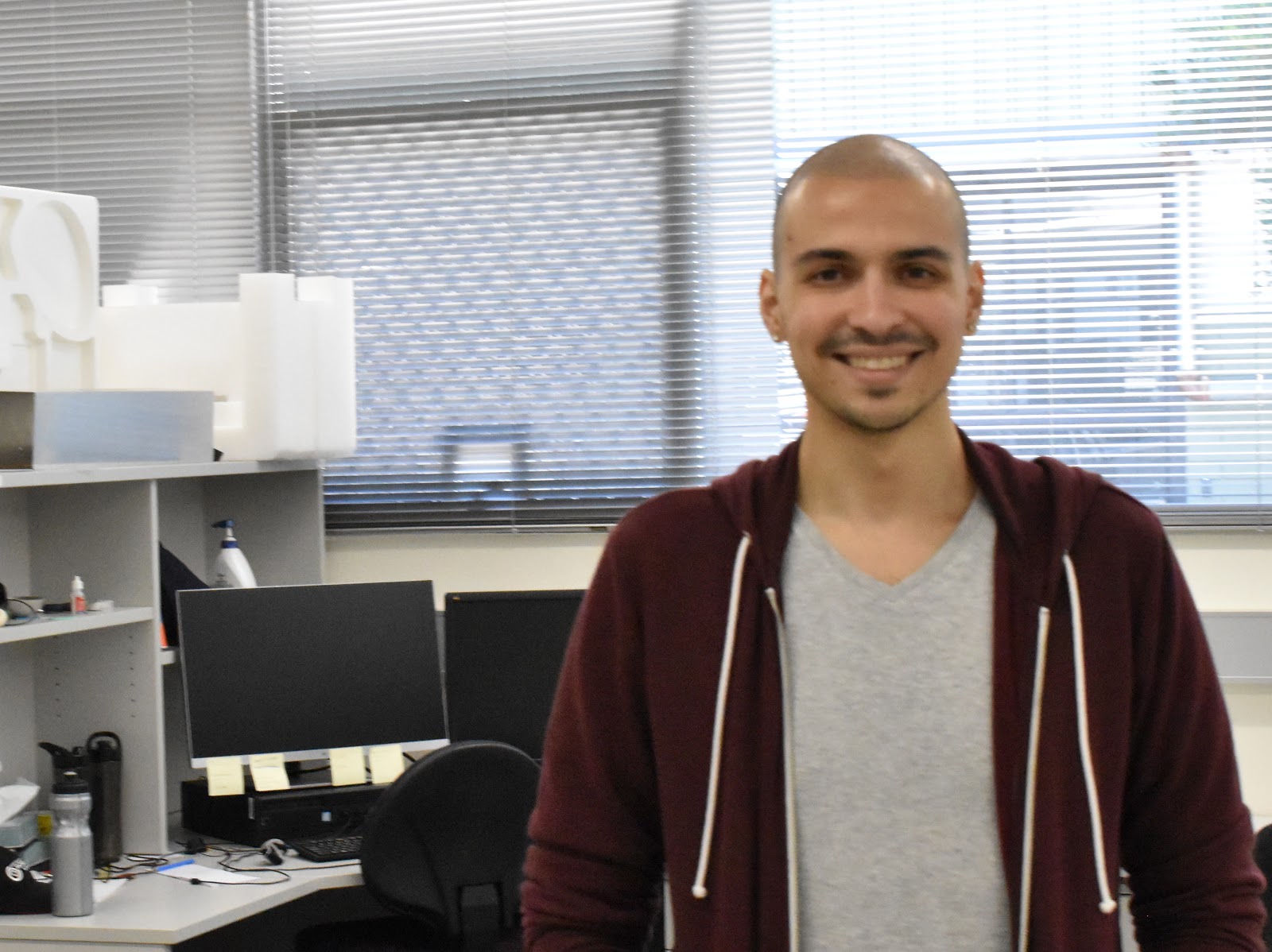}} \hspace{1pt}
    \subfloat[Mean rating: $4.2$]{\includegraphics[width=0.32\linewidth]{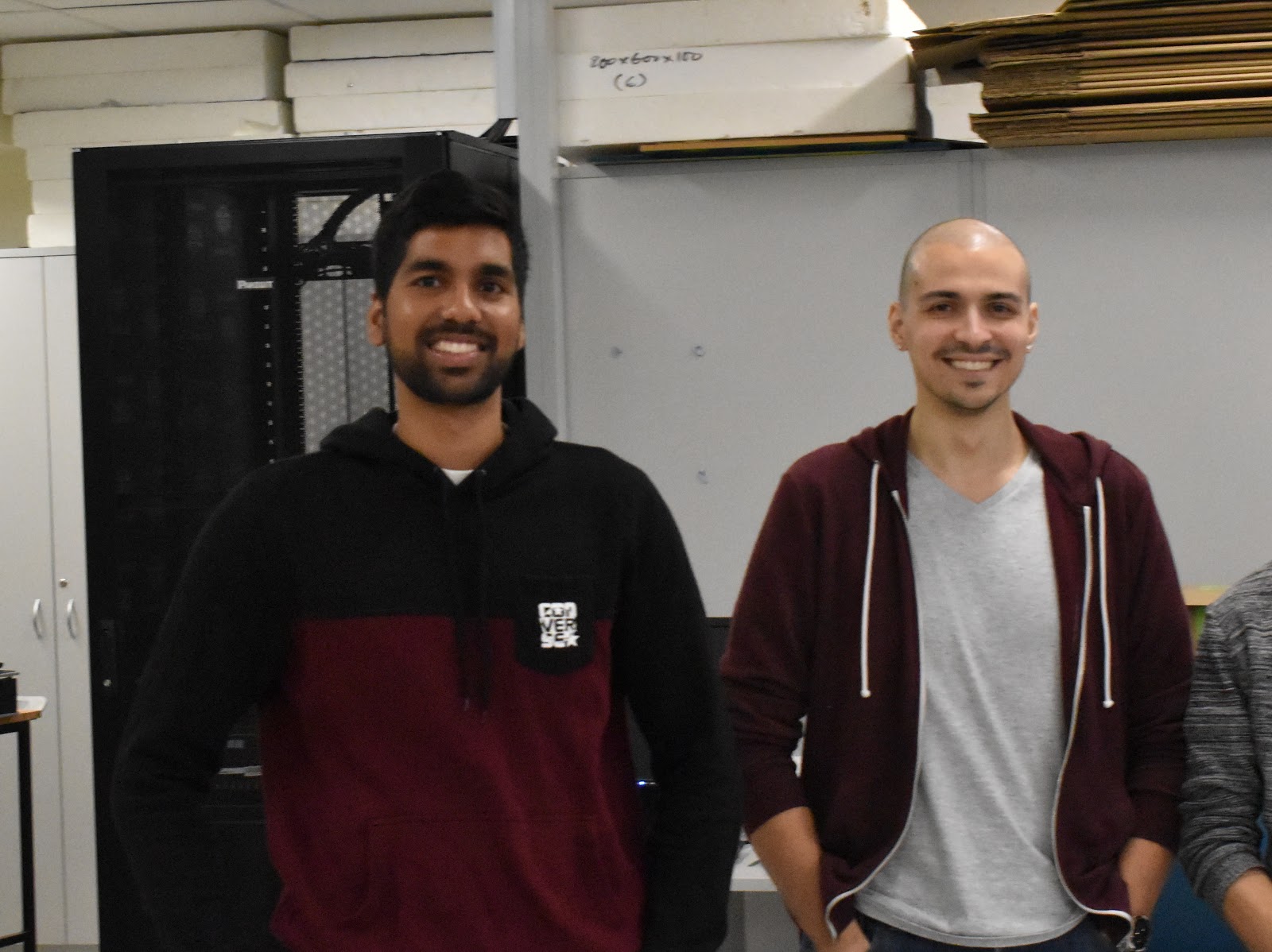}} \hspace{1pt}
    \subfloat[Mean rating: $4.12$]{\includegraphics[width=0.32\linewidth]{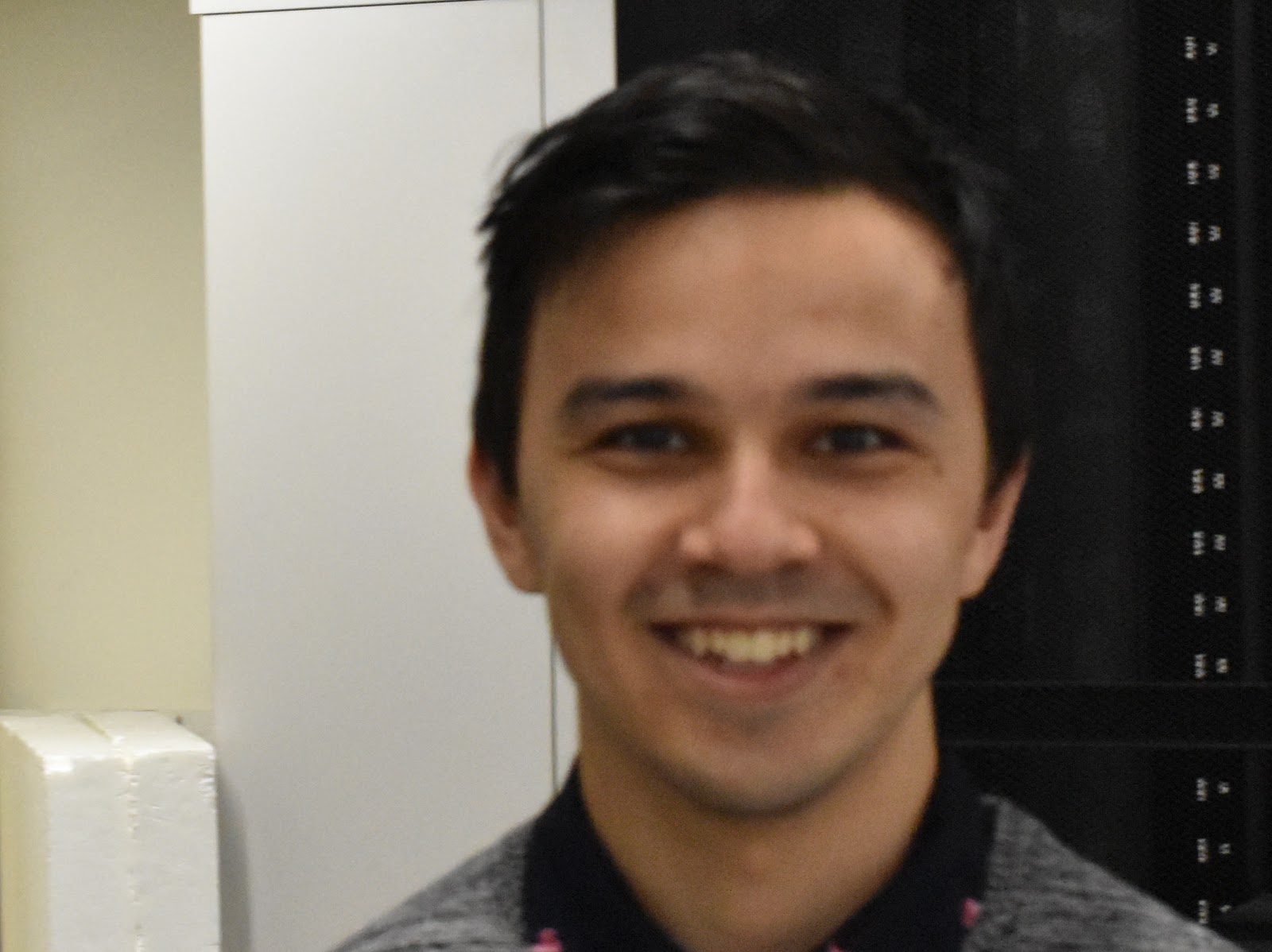}} \hspace{1pt}
    \\
    \subfloat[Mean rating: $3.84$]{\includegraphics[width=0.32\linewidth]{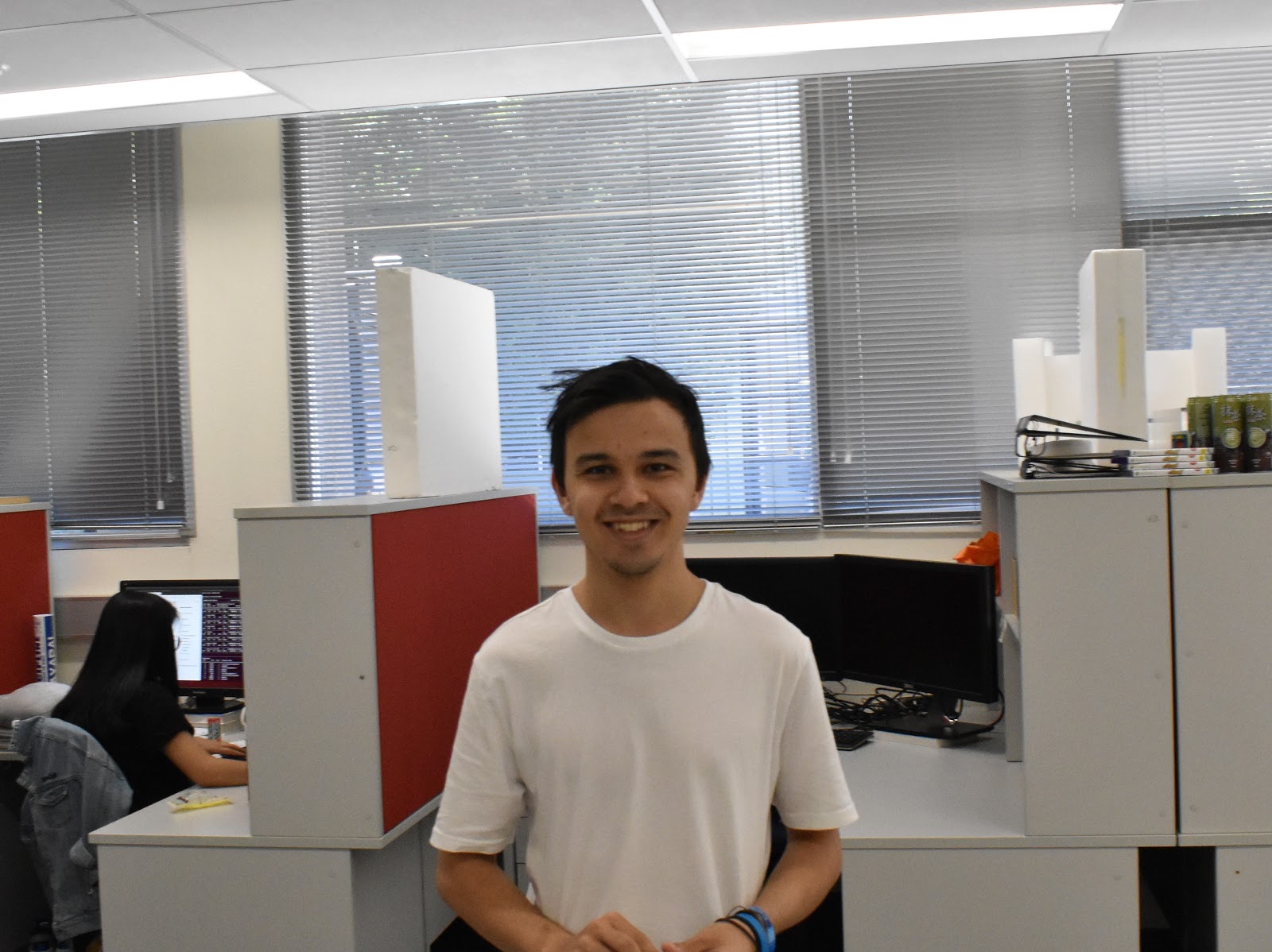}} \hspace{1pt}
    \subfloat[Mean rating: $3.78$]{\includegraphics[width=0.32\linewidth]{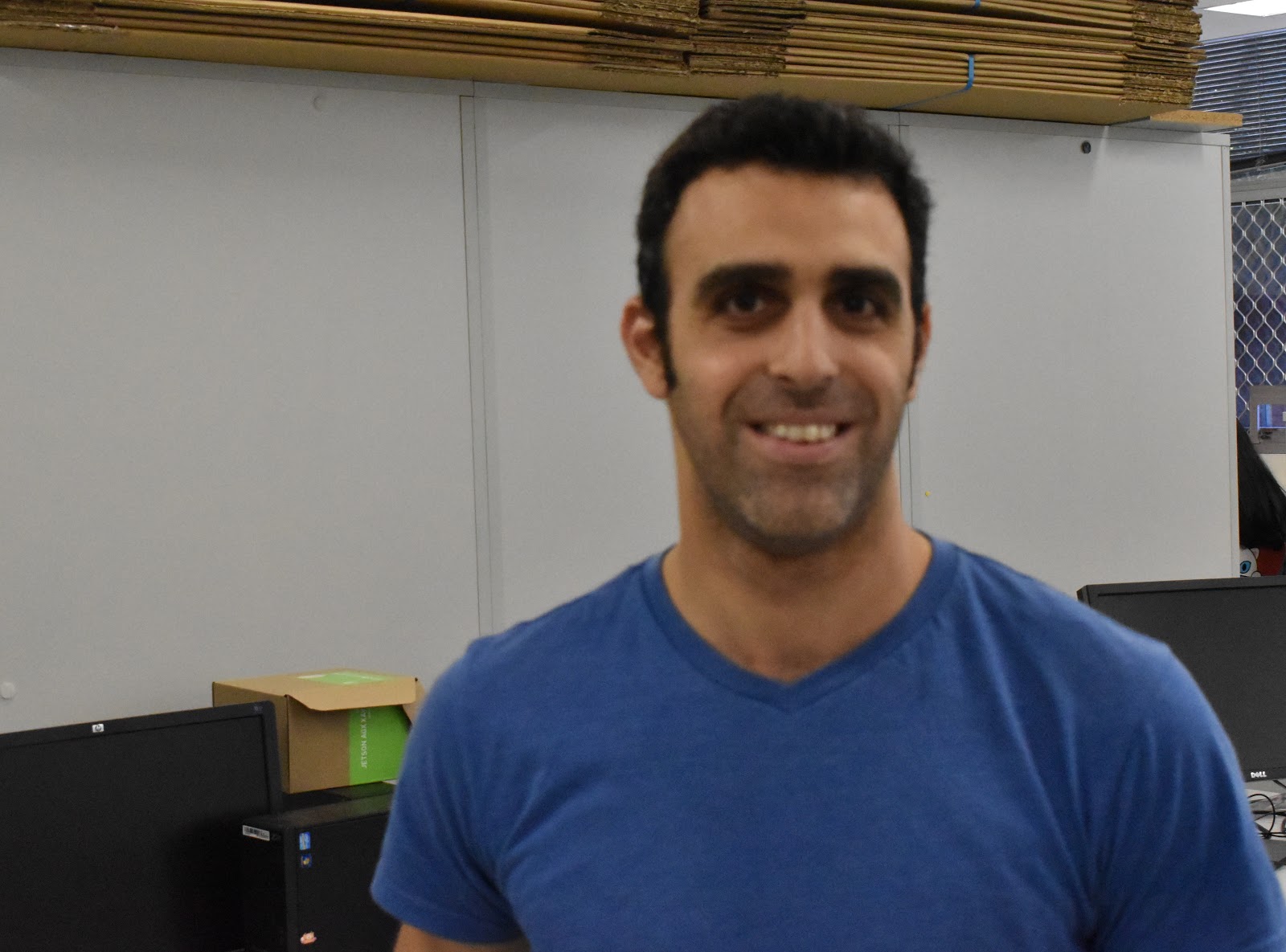}} \hspace{1pt}
    \subfloat[Mean rating: $3.75$]{\includegraphics[width=0.32\linewidth]{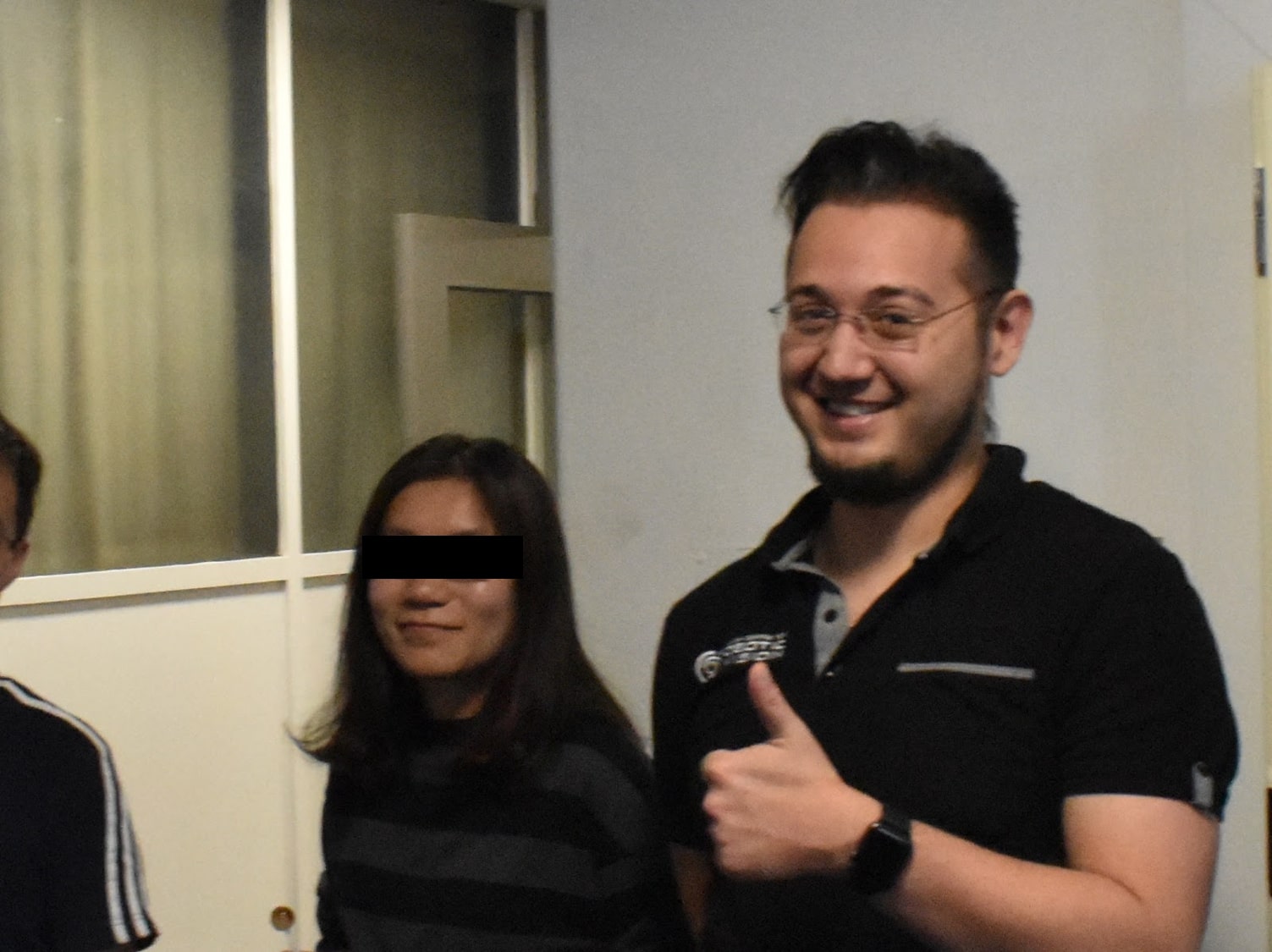}} \hspace{1pt}
    \\
    \subfloat[Mean rating: $3.08$]{\includegraphics[width=0.32\linewidth]{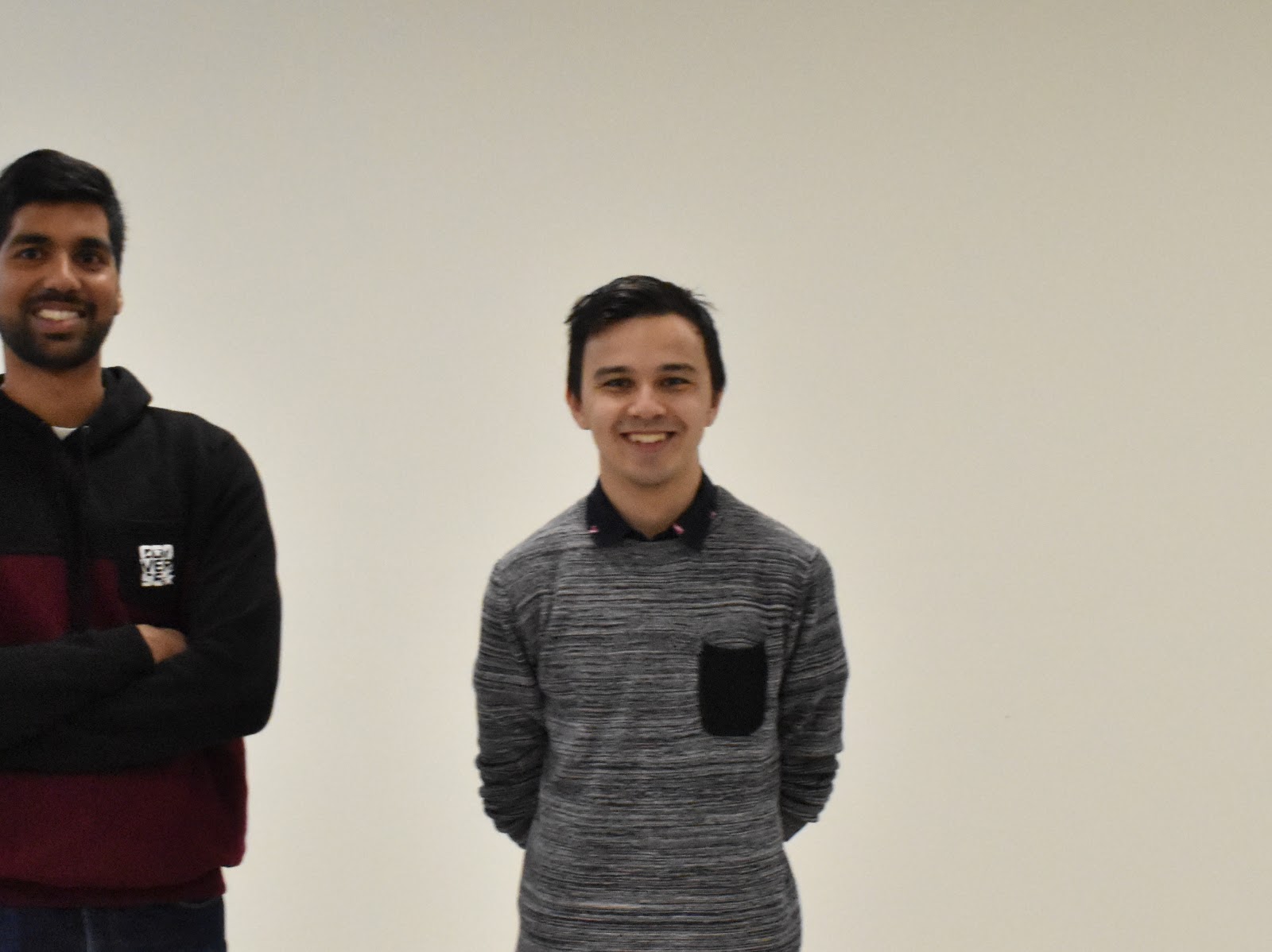}} \hspace{1pt}
    \subfloat[Mean rating: $2.76$]{\includegraphics[width=0.32\linewidth]{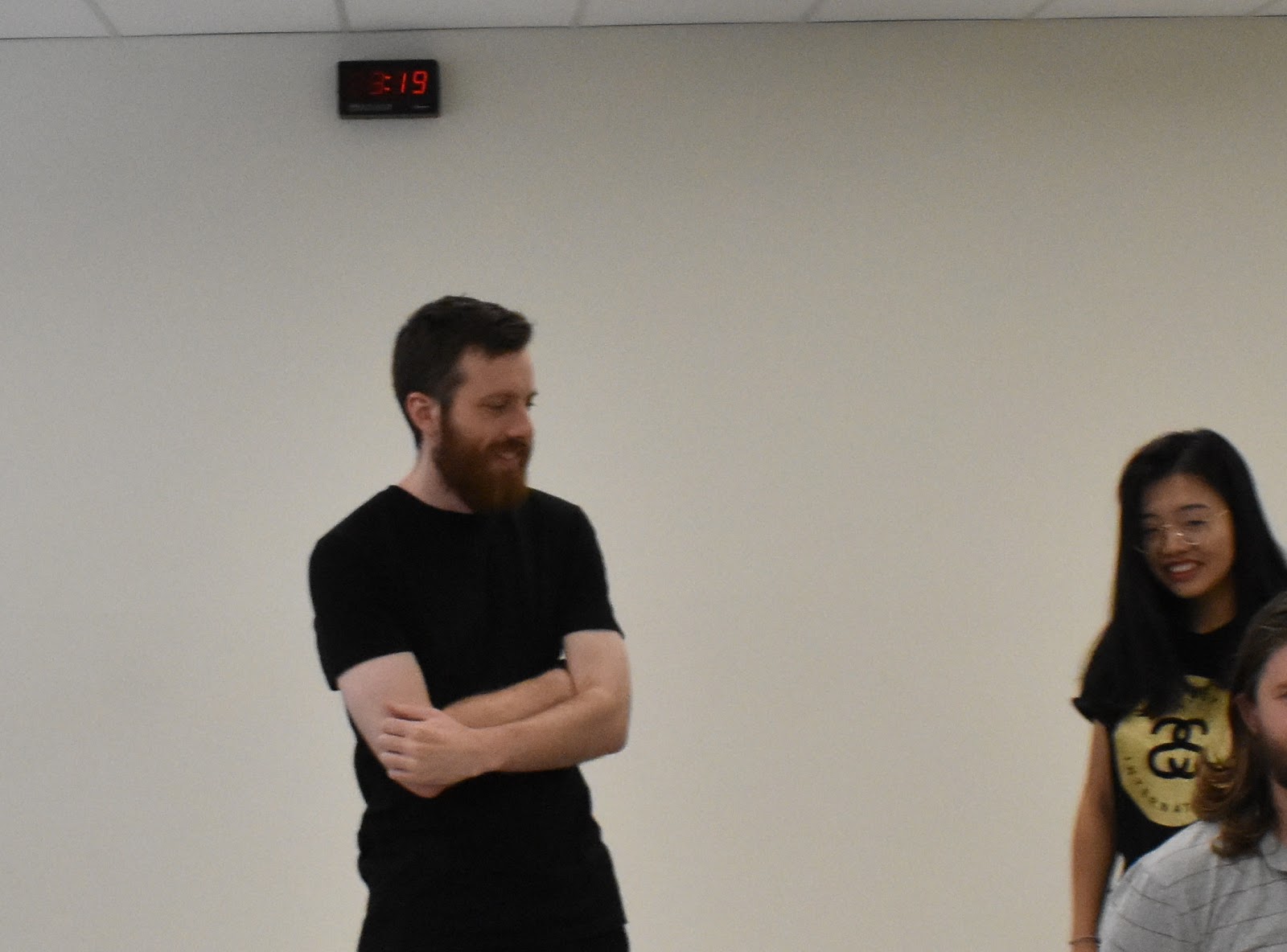}} \hspace{1pt}
    \subfloat[Mean rating: $2.72$]{\includegraphics[width=0.32\linewidth]{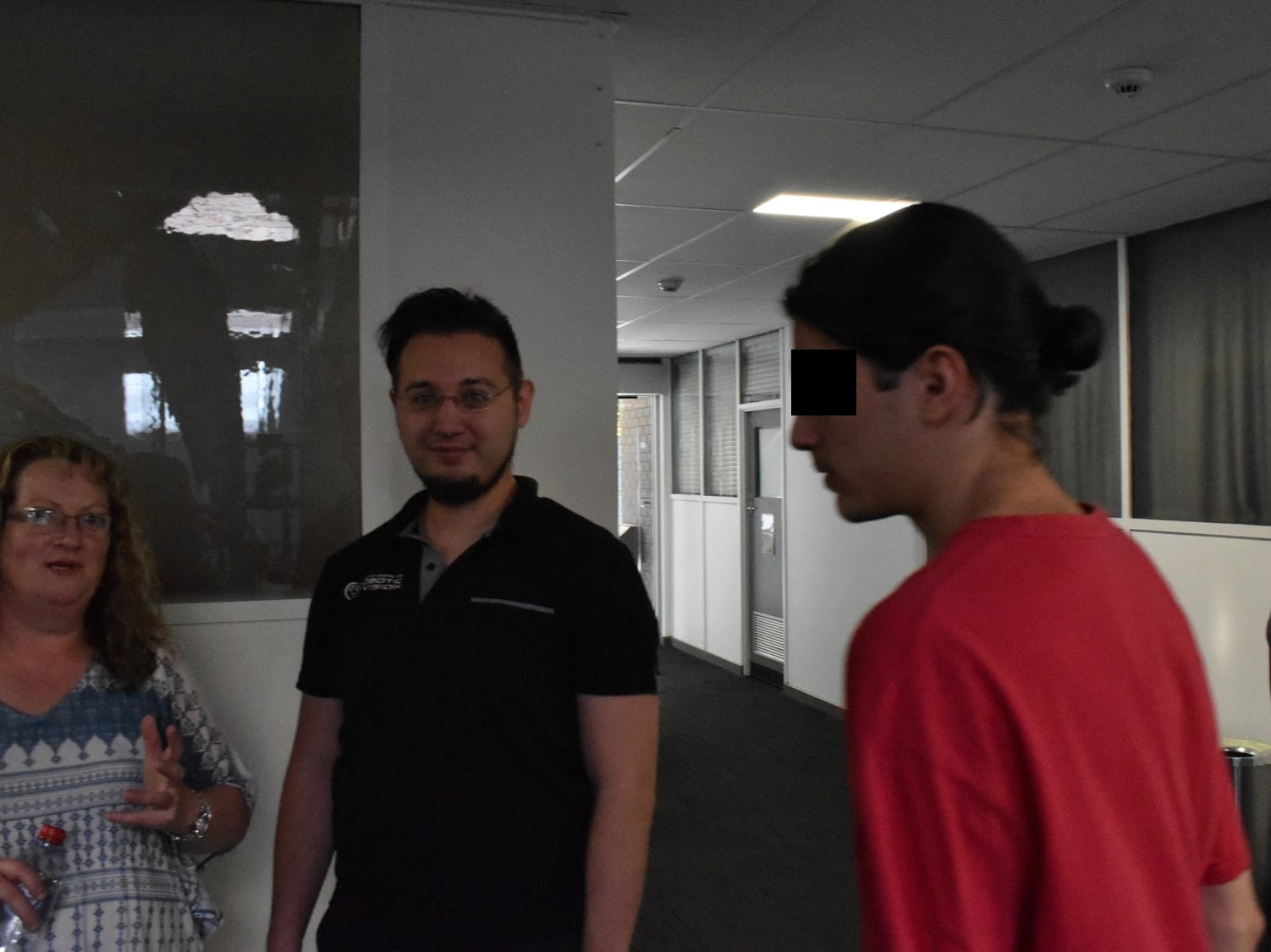}} \hspace{1pt}
    \caption{Example pictures that were selected as the best using the Picture CNN approach, along with the mean ratings from the user study. Top Row: Top 3 rated pictures. Mid Row: Pictures around the median rating. Bottom Row: Worst 3 rated pictures.}%
    \label{fig:user_study_qualitative}%
\end{figure}

We conducted two t-tests~\cite{student1908probable} to test \textit{H1}. The null hypothesis that claims there is no difference between CNN and Heuristic rating distributions was rejected with \(p=0.0025\) ($T=2.81$). Null hypothesis that claims there is no difference between CNN and Baseline ratings was rejected with even stronger evidence $p=0.0001$ ($T=4.25$). Therefore, we can say with confidence that CNN outperformed both Baseline and Heuristic, confirming Hypothesis 1.

\begin{figure}[h!]
\centering
\includegraphics[clip, trim=0.25cm 0.3cm 0.25cm 0.25cm, width=0.5\textwidth]{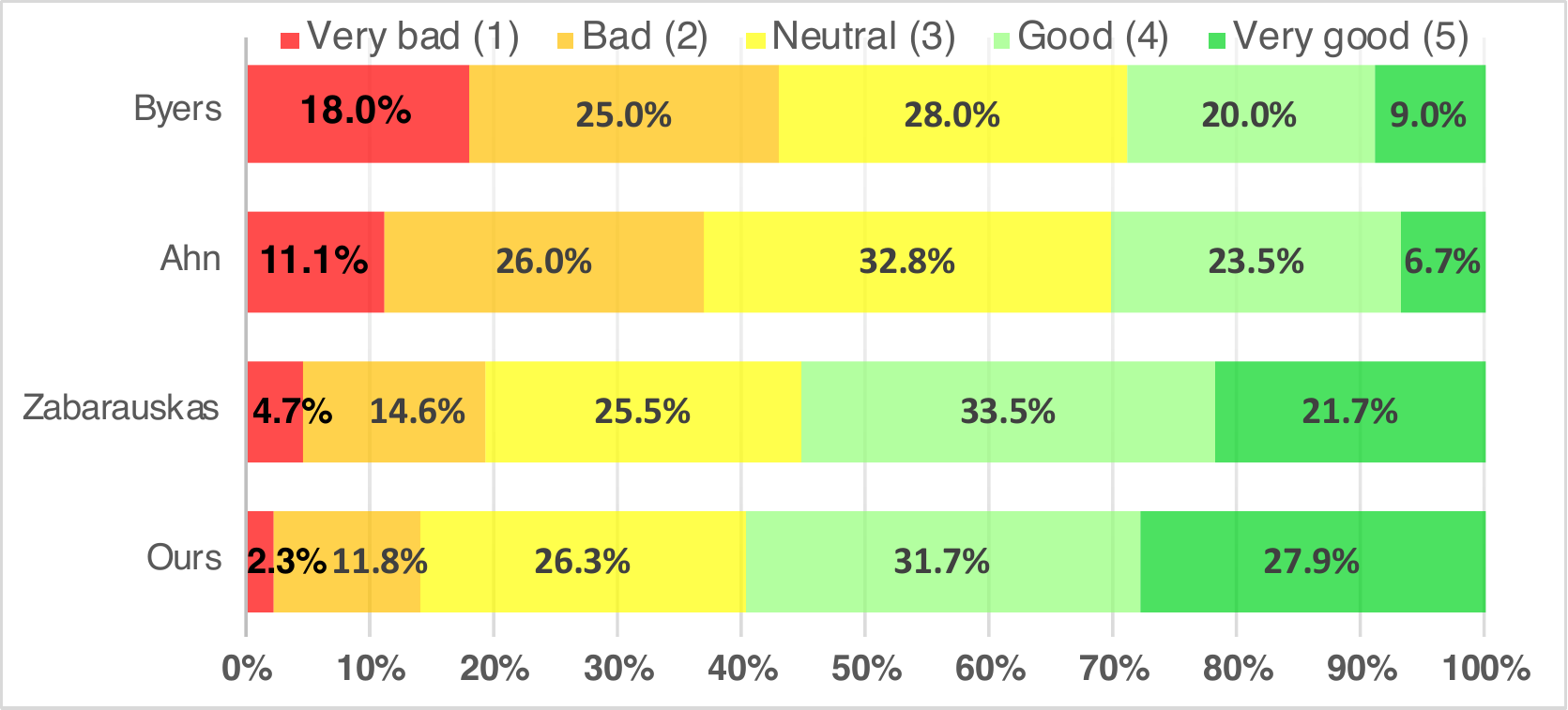}
\caption{Proportion of pictures in each rating category are shown in comparison with the state of the art}
\label{fig:comparison}
\end{figure}

We also compared the ratings obtained with the CNN method with others in the literature~\cite{zabarauskas2014luke,Ahn,Byers}. Due to difference in robot platforms, we could not implement their algorithms. Instead, we compare the user rating distributions as a pseudo metric for comparison, even though the pictures and subjects were different. The rating distributions are compared with the state of the art in Figure \ref{fig:comparison}. Please note that Ahn et al.~\cite{Ahn} used the following Likert Scale: Very Poor, Poor, Normal, Nice and Very Nice.

\begin{table}[h!]
\centering
\begin{tabular}{l|c|c|l}
Method    & $\mu$  & $\sigma$ & Num. Responses\\ \hline
Byers~\cite{Byers}  & 2.77 & 1.22 & 2000$^*$\\
Ahn~\cite{Ahn} & 3.11 & 1.09 & 1040\\
Zabarauskas~\cite{zabarauskas2014luke} & 3.53 & 1.12 & 1648\\
Ours (Picture CNN) & 3.71 & 1.06 & 2328\\
\end{tabular}
\caption{Comparison of user study results with the state of the art. $^*$Byers et.al. mentions over 2000 pictures were evaluated.}
\label{tab:user_study_state_of_the_art}
\end{table}

Table \ref{tab:user_study_state_of_the_art} shows the comparison with the state of the art. Pictures chosen by our CNN method received \textit{(5) Very good} rating $27.9\%$ of the time, which is higher than others. The highest proportion after ours is Zabarauskas~\cite{zabarauskas2014luke} with $21.7\%$. Our method has the highest mean rating among all. A series of t-tests showed that the difference from other results is statistically significant, with p-values virtually zero. This confirms \textit{H2}. Our approach in also the most consistent in taking pictures, having the lowest standard deviation out of all methods.

\section{Conclusion and Future Work}
\label{sec:conclusion}
 
In this paper we describe a deep learning based system to perform robot photography. The robot follows a line defined by the end-user and takes a series of burst photos of subjects. These pictures are classified as either \textit{Good} or \textit{Bad} based on quality of faces and their relative positioning. The quality of face images are estimated using features such as the face orientation and likelihood of facial emotions. A user study was conducted where 97 human judges rated the pictures taken by the robot photographer. The results suggest a statistically significant improvement over a heuristic method based on hand-crafted photographic composition rules and other published work. 

Two potential areas of improvement are robot navigation and human-robot interaction (HRI). Currently the robot is constrained to the line and and its behavior is not actively guided by the quality of the images. The robot can instead explore the environment and use an objective function to search for a good picture. Better HRI would likely boost the quality of the images, for example, if the robot asks interested people to pose and smile for the camera.










\section{Acknowledgment}
\label{sec:acknowledgment}
\small We would like to thank Aryaman Pandav, Augustus Hebblewhite and Haluk Koseoglu for their help with various parts of the project.

\begin{spacing}{0.9}
\bibliographystyle{is-unsrt}
\bibliography{refs}
\end{spacing}


\end{document}